\pdfoutput=1
\documentclass[10pt,twocolumn,letterpaper]{article}

\pdfoutput=1
\usepackage{cvpr}
\usepackage{times}
\usepackage{graphicx}
\usepackage{amsmath}
\usepackage{amssymb}
\usepackage{caption}
\usepackage[nocompress]{cite}
\usepackage{slashbox}
\usepackage[dvipsnames,table]{xcolor} 
\usepackage[toc,page]{appendix}
\usepackage{enumerate}
\usepackage{tabularx,ragged2e}
\usepackage{spreadtab}
\usepackage{adjustbox}
\usepackage{booktabs}
\usepackage{xcolor}
\usepackage{multirow}
\usepackage{rotating}
\usepackage{color, colortbl}
\usepackage{subfigure}
\usepackage{xcolor}
\usepackage{soul}

\usepackage[inline]{enumitem}
\usepackage{arydshln}

\usepackage{authblk}

\graphicspath{{./figures/}}

\DeclareGraphicsExtensions{.jpeg,.png,.eps,.pdf}

\definecolor{Gray}{gray}{0.9}

\definecolor{Gray}{gray}{0.85}
\definecolor{LightCyan}{rgb}{0.88,1,1}

\newcolumntype{a}{>{\columncolor{Gray}}c}
\newcolumntype{b}{>{\columncolor{white}}c}

\newcommand{\partitle}[1]{\bigbreak\noindent\textbf{#1}}

\makeatletter
\newcommand*{\rom}[1]{\expandafter\@slowromancap\romannumeral #1@}
\makeatother

\cvprfinalcopy 

\def\cvprPaperID{4} 

\ifcvprfinal\fi

\begin{document}


\title{Continual Learning of Object Instances}




\author[1,2]{Kishan Parshotam}
\author[1]{Mert Kilickaya}

\affil[1]{University of Amsterdam, The Netherlands}
\affil[2]{Prosus, The Netherlands}

\affil[ ]{\textit {kishan.parshotam@\{student.uva.nl, prosus.com\}}}
\affil[ ]{\textit {kilickayamert@gmail.com}}

\maketitle
\thispagestyle{empty}

\begin{abstract}

We propose continual instance learning - a method that applies the concept of continual learning to the task of distinguishing instances of the same object category. We specifically focus on the car object, and incrementally learn to distinguish car instances from each other with metric learning. We begin our paper by evaluating current techniques. Establishing that catastrophic forgetting is evident in existing methods, we then propose two remedies. Firstly, we regularise metric learning via Normalised Cross-Entropy. Secondly, we augment existing models with synthetic data transfer. Our extensive experiments on three large-scale datasets, using two different architectures for five different continual learning methods, reveal that Normalised cross-entropy and synthetic transfer leads to less forgetting in existing techniques.


\end{abstract}

\section{Introduction}
\begin{figure}[t]
\vspace{0.5cm}
\hbox{\hspace{2.0em}\includegraphics[scale=0.65]{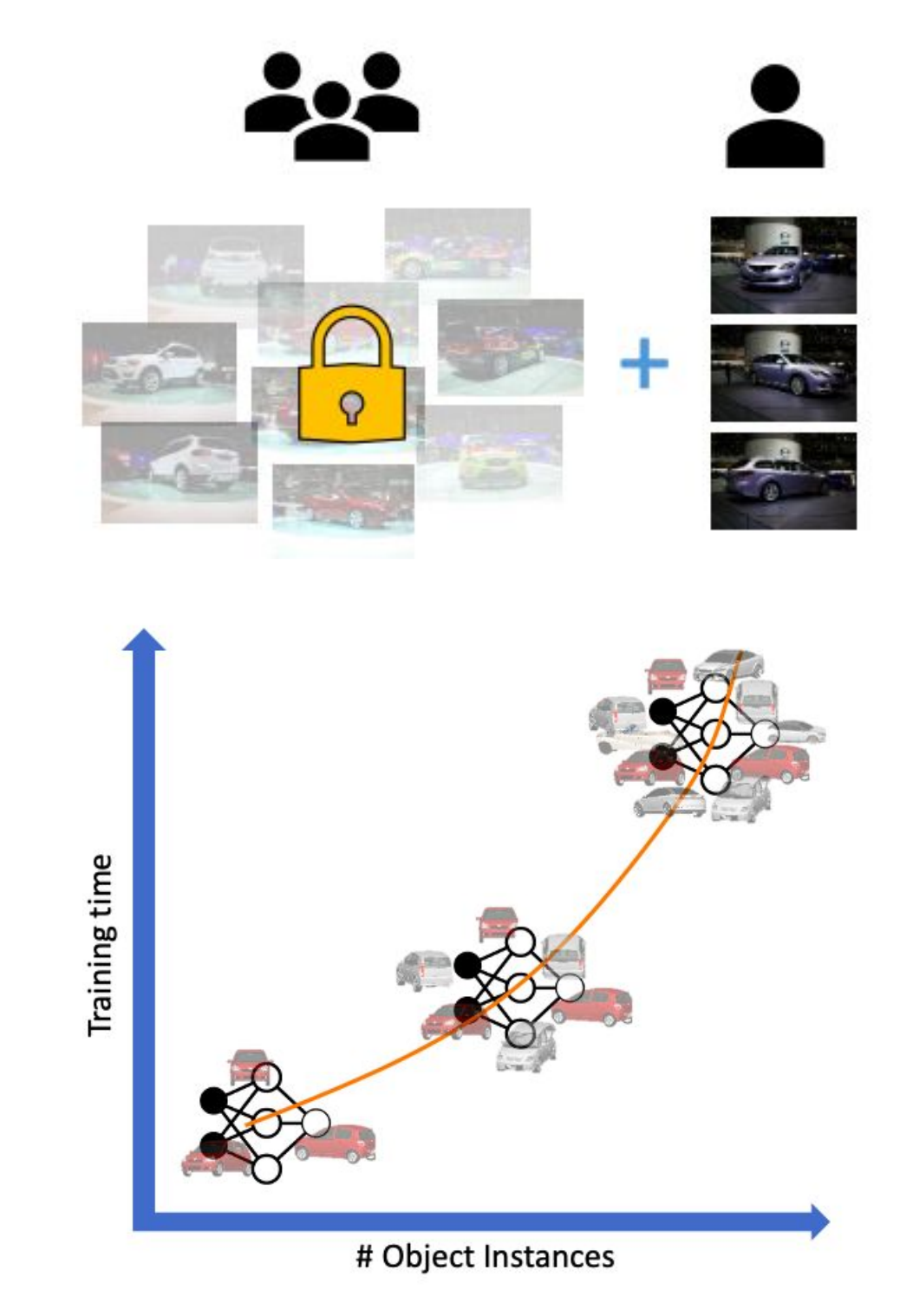}}
  \caption{Two cases for continual learning of object instances. (Top) Retail companies receive thousands of novel car instances daily. However, these are not always accessible due to privacy reasons. (Bottom) Surveillance companies receive long streams of car records, therefore training from scratch becomes inefficient. In this paper, we propose continual learning of object instances to tackle such issues of privacy and inefficiency.}

 
  \label{fig:teaser}
\end{figure}

\label{section:Intro}

Most computer vision tasks assume access to a full, static dataset. Regularly, researchers train and test their algorithms on well-established benchmarks~\cite{imagenet,pascal,mscoco}. Although beneficial from a benchmarking point of view, this setup neglects the dynamic, ever-changing nature of the visual world. The world does not present itself as a static set of objects that remain similar through time. Humans evolved to be life-long learners \cite{mclelland} and update their visual model of the world with sensory data. Therefore, continual learning (CL) is proposed to mimic learning about novel object categories~\cite{lwf, lfl, kirkpatrick2017overcoming}.


 A common task for CL is to learn about a new object (\eg chair) while retaining performance on the previously seen (\eg car, table, bicycle). In this work, we focus on \textit{continuously} learning to distinguish between different instances of the same object category, in our case, cars. Continual instance learning (CIL) is an approach that applies the concept of CL to the task of distinguishing intra-category instances through \textit{metric learning}~\cite{VehicleReID_Triplet,facenet,face_veri}.

A continuous stream of car objects is evident in the visual world, especially for cases of retail and surveillance, consider Figure~\ref{fig:teaser}. For instance, an online car retail company receives numerous new car advertisements daily. Ideally, a company would continuously learn from the incoming source of car images, for applications such as identifying duplicate ads~\cite{duplicate}. However, due to privacy reasons, companies may need to delete historical user records, which affects model (re-)training \cite{empiricalCatastrophicForgetting, core50}. Likewise, a surveillance company may aim to learn to re-identify a query vehicle within the city. However, learning vehicle re-identification from scratch is not optimal, since the amount of car images is ever-increasing. We address such scenarios with continual instance learning.

Our contributions are as follows:
\begin{enumerate}
    \item We introduce the problem of \textit{continual instance learning} by evaluating existing continual learning methods under the metric learning setup. We show that existing techniques suffer from \textit{catastrophic forgetting}.
    \item We propose to utilise Normalised Cross-Entropy (NCE)~\cite{sim_clr}, to reduce the effects of catastrophic forgetting by mitigating the sensitivity to outliers under regression losses \cite{InDefenseOfTripletLoss,facenet}. 
    \item We augment existing techniques with synthetic visual data transfer, showing improvements on three benchmark datasets over two backbone architectures.
\end{enumerate}

\section{Related Work}


\subsection{Continual Learning}


Under different training scenarios, neural networks suffer from catastrophic forgetting~\cite{McCloskey_Cohen_1989, empiricalCatastrophicForgetting, core50}. This pitfall has inspired the research community to reestablish neural networks as a system that is not only put on hold once trained but as one that can improve through time. This has given rise to continual learning, also known as incremental learning, formally introduced in~\cite{core50}. In continual learning, one does not have access to previously seen data.


Our work follows more closely the New Instances (NI) setting \cite{core50} - new training data of the same classes become available with new poses and conditions. We differentiate ourselves by presenting new instances of only a single class object and learn to better distinguish these. We focus on regularisation approaches since these have shown to have adequate performances with a low level of complexity \cite{lfl, lwf, kirkpatrick2017overcoming}. Early attempts regularise the loss function by maintaining the output of the network as unchanged as possible while shifting the internal feature representation, namely Less Forgetting Learning (LFL) ~\cite{lfl} and Learning without Forgetting (LwF) ~\cite{lwf}. In Elastic Weight Consolidation (EWC) importance is defined for each model parameter via a Bayesian approach \cite{kirkpatrick2017overcoming}.
This research is closely related to \cite{core50} and \cite{incremental_learning} where we differentiate ourselves from the classification setup and introduce metric learning to continual learning.

\subsection{Instance Learning}


In metric learning, we are directly learning a distance function between objects in a $D$-dimensional space. Applications in this domain extend to image retrieval tasks, specifically, face verification \cite{face_veri}, person and vehicle re-identification (ReID)~\cite{facenet, VehicleReID_Triplet}. In this work, we focus on the vehicle re-identification problem. Common vehicle ReID learning approaches resort to Siamese CNN \cite{siameseCNN} with contrastive \cite{contrastive_loss} or triplet losses \cite{facenet, VehicleReID_Triplet}. These introduce an effective method of separating objects in the embedding space.

Directly translating CL approaches to metric learning is not possible due to the different problem formulation - these approaches build upon regularising classification outputs, whereas metric learning directly attempts to learn a manifold. This approach is susceptible to outliers~\cite{facenet} due to the unbounded nature of the gradients. To this end, we propose Normalised Cross-Entropy~\cite{sim_clr} as a solution. More specifically, we treat metric learning as a classification task, where given a query image, we solve a binary task to estimate whether a retrieved car image belongs to the same query under a different viewpoint. 

\subsection{Transfer Learning from Synthetic Data}
The successful effect of using synthetic data for improving machine learning systems has been reported in \cite{syntheticData}. It reduces the need for labels \cite{less_labels} and has shown to help in different visual tasks \cite{humanpose_synthetic, synthetic_fog}. To reduce forgetting, we evaluate the effect of pre-training a model on object-specific discriminant features. For this, we propose to utilise synthetic transfer learning for CIL.

\smallskip


\section{Approach}

\label{section:CLM}


\subsection{Continual Learning}

Regularisation CL approaches are easily adapted to systems in production. These do not rely on historical samples and do not increase the number of model parameters.
In our experiments, we focus on Less Forgetting Learning~\cite{lfl}, Learning without Forgetting~\cite{lwf} and Elastic Weight Consolidation~\cite{kirkpatrick2017overcoming}. The first two approaches focus on keeping the decision boundary as unchanged as possible, whereas the latter targets updating internal representations. For comparison, we also run our experiments on baseline CL approaches, specifically, Naïve and Fine-tuning.

We define $X_o$ and $Y_o$ as the input and targets of an \textit{old} dataset that is no longer available. Similarly, for a \textit{new} dataset, we define $X_n$ and $Y_n$. Additionally, the parameters of a previously trained model and the new model are defined as $\theta_o$ and $\theta_n$ respectively.

\partitle{Naïve:}
The benchmark approach for continual learning is to simply re-train a model on $X_n$ and $Y_n$ given $\theta_o$. This training procedure does not require any additional parameters and by definition, it yields $\theta_n$.

\partitle{Fine-Tuning (FT):}
One simple approach to bypass catastrophic forgetting is to first train a model on $X_o$ and $Y_o$, and once we obtain access to $X_n$ and $Y_n$, train a model by fine-tuning discriminant layers \cite{lwf}.

\partitle{Less Forgetting Learning (LFL) \cite{lfl}:}
One important characteristic in CL is to ensure that we get similar predictions $\hat{Y}$ for $X_n$ under $\theta_o$ and $\theta_n$. To this end, in \cite{lfl} the authors propose to improve the re-training process by initialising a model with $\theta_n = \theta_o$. Additionally, it is proposed to freeze the softmax layer so that the decision boundaries of the model remain similar. The authors train such a model with the following loss function:

\begin{equation}
\label{eqn:lfl}
\mathcal{L}_{LFL}(x_n; \theta_o; \theta_n) = \lambda_c \mathcal{L}_c(x_n;\theta_n) + \lambda_e \mathcal{L}_e(x_n;\theta_o;\theta_n),
\end{equation}
where $\mathcal{L}_c$ and $\mathcal{L}_e$ are the cross-entropy and the Euclidean loss functions. The authors constrain feature changes with the Euclidean loss between the predictions of $X_o$ under $\theta_n$ and $X_o$ under $\theta_o$.

\partitle{Learning without Forgetting (LwF) \cite{lwf}:}
Similarly, in LwF \cite{lwf}, the authors encourage the network to keep its features as unchanged as possible by using the Knowledge Distillation loss \cite{kd} as opposed to the Euclidean loss in Equation \ref{eqn:lfl}. However, in this approach, $\theta_n$ is randomly initialised and there is no explicit network freezing.

\partitle{Elastic Weight Consolidation (EWC) \cite{kirkpatrick2017overcoming}:}
Restricting feature extraction capabilities is not optimal since we want a model to learn from new instances and incrementally improve its object representation. Therefore, in~\cite{kirkpatrick2017overcoming} the model parameters $\theta$ take a probabilistic form with the Fisher Information matrix, $F_i$, reflecting weight importance. Under CL, the model is regularised with the following loss function:

\begin{equation}
\label{eqn:ewc}
    \mathcal{L}_{EWC} =  \mathcal{L}_n + \sum_i \frac{\lambda}{2}F_i (\theta_{n,i} - \theta_{o,i}),
\end{equation}{}

where $\mathcal{L}_n$ is the loss function for the new task, and $\lambda$ is the previous task importance.

\subsection{Regression-based Metric Learning}
We train the vehicle ReID datasets using a siamese network and the triplet loss \cite{facenet}. We distance dissimilar vehicles and cluster similar ones in the manifold space. Specifically, we define an anchor ($x_{i}^a$), a positive ($x_{i}^p$) and a negative sample ($x_{i}^n$). The loss function is defined as:
\begin{equation}
\label{eqn:triplet}
    \sum_{i}^{N}\left[\left\|f\left(x_{i}^{a}\right)-f\left(x_{i}^{p}\right)\right\|_{2}^{2}-\left\|f\left(x_{i}^{a}\right)-f\left(x_{i}^{n}\right)\right\|_{2}^{2}+\alpha\right]_{+},
\end{equation}{}where $\alpha$ is defined as the margin which, similar to \cite{InDefenseOfTripletLoss}, is set to 1. \textit{Mining} hard negative pairs is highly important to obtain discriminant features \ie we want to emphasise training of instances which are harder to distinguish, such as two different cars of similar colour and pose, as opposed to trivially distinct cars.

\section{Continual Instance Learning}


\subsection{Continuous Batches}
\label{section:datasplit}
We define three data streaming approaches for accessing object instances under CIL, where one time-step, or a new training set, is a \textit{continuous batch}. Each approach presents different restrictions and is suitable for different applications.
\begin{enumerate}
    \item \textbf{Random:} Upon acquiring new data, the particular object instances in the set are not known \textit{a priori}. Therefore, this approach assumes that in each continuous batch there can be different instances and each of them can have a different number of data-points. Under the triplet loss, we do not know the number of negative instances, or how hard these are to learn from.
    \item \textbf{Balanced:} Some tasks can also acquire new instance data in a balanced fashion \ie the same number of data-points are collected in each time step for all instances. If this is the case, we hypothesize that it is easier to mine for hard negatives, since we have access to more negative instances.
    \item \textbf{Incremental:} On the other hand, some applications acquire only some instances at a time \ie every continuous batch is restricted to only a few instances. This approach has limited access to negative instances and is, therefore, the most challenging.
\end{enumerate}{}
In our experiments, we assume the most challenging scenario, incremental continuous batches, which is also most closely related to the examples portrayed in section \ref{section:Intro}.

\subsection{Adapting current approaches}
We adapt CL approaches to CIL for bench-marking purposes. To achieve this, we propose the following changes:

\begin{itemize}
    \item For the Naïve and Fine-Tuning approaches, we directly replace the cross-entropy loss \cite{lwf} with the triplet loss.
    \item For LFL and EWC we propose the same approach by replacing $\mathcal{L}_c$ from Equation \ref{eqn:lfl} and  $\mathcal{L}_n$ from Equation \ref{eqn:ewc} with the triplet loss.
    \item In the LwF scenario, the authors make use of the knowledge distillation loss, which encourages probabilities of one network to approximate the output of another network. Since in the regression-based triplet loss we are not dealing with probabilities, we are therefore not able to adapt this approach.
\end{itemize}{}

\subsection{Normalised Cross-Entropy Loss (NCE)}
Two unwanted features in continual instance learning are, 1) dealing with a regression setup which is not robust to outliers and 2) not being able to build upon current CL approaches, like LwF.

Having an outliers-robust approach allows us to have instances that are extremely different from all others and still restrict its gradients. This can be a fairly common scenario in incremental continuous batches. To this end, the triplet loss is adapted to make use of the normalised cross-entropy loss \cite{sim_clr},

\begin{equation}
    \label{eqn:nce}
    \ell({z}_i)= - \log \frac{\exp(z_{i} / \tau)} {\sum_{j=1}^{K} \exp(z_{j} / \tau)} \text { for } i=1, \ldots, K+1
\end{equation}{}

where, 

\begin{equation}
    \mathbf{z}=\left({x^{a}}^\top x^{p}, {x^{a}}^\top x^{n}_1 \ldots, {x^{a}}^\top x^{n}_K\right).
\end{equation}{}

We set the temperature, $\tau = 1$ for simplicity. We do not resort to offline or online negative mining since we are constrained by the number of hard negative samples. However, we allow for the loss function to include more than one negative sample at a time ~\cite{NML}. We define the target of $\mathbf{z}$ as 1 for the index corresponding to the dot product between the anchor and the positive pair and 0 for all other pairs

\subsection{Synthetic Visual Data Transfer}
We first train a model on synthetic instances of the same object as the real object of interest. We then apply CIL under new real instances. We motivate this approach to ensure that the model first fully learns fine-grained visual representations from the synthetic data before training on continuous batches, which may be of skewed distributions.


\section{Experimental Setup}


\begin{figure}[h]
  \hbox{\hspace{-1.1em}\includegraphics[scale=0.55]{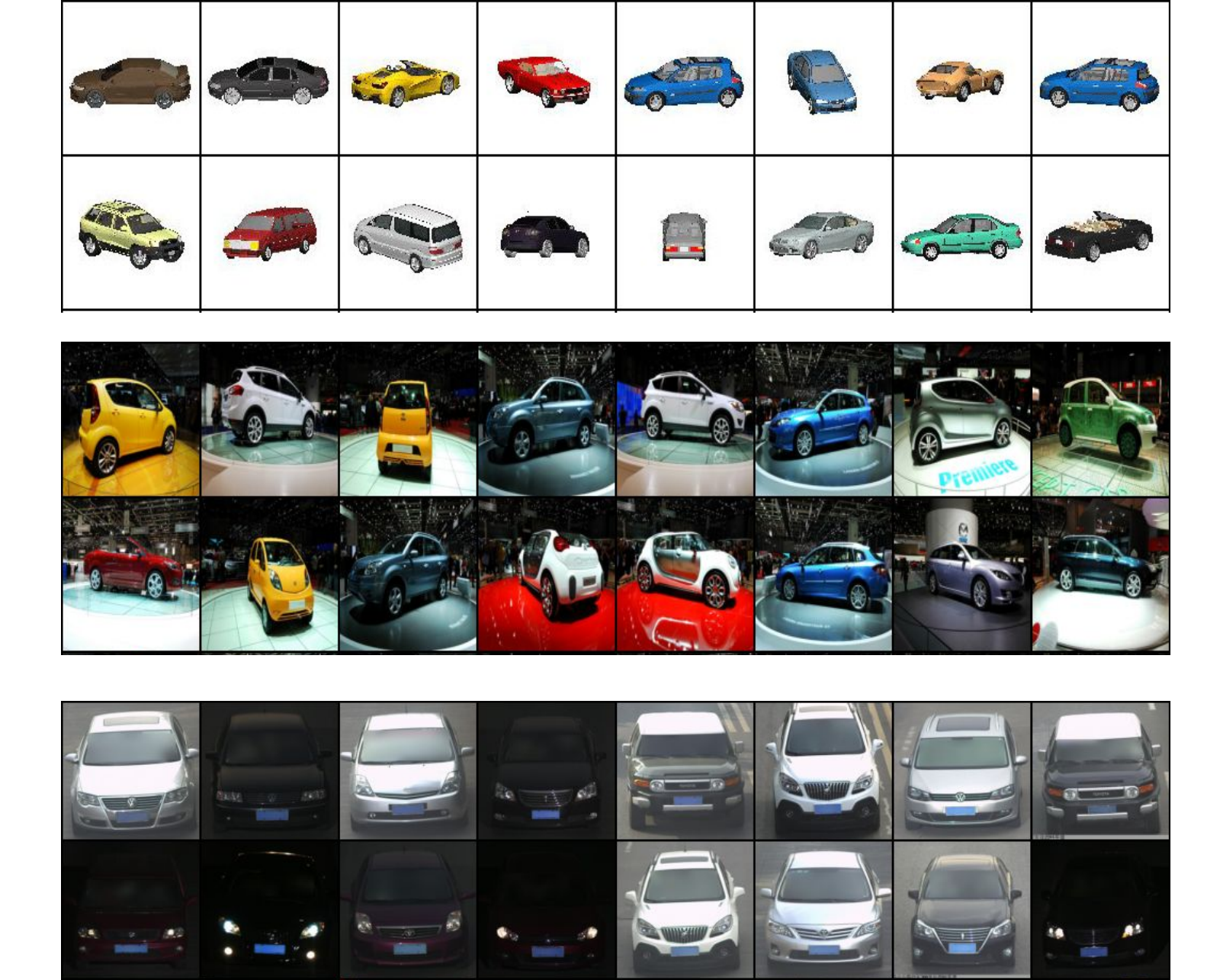}}
  \caption{Example images from the three datasets, namely Cars3D (top), MVCD (middle) and CompCars (bottom). Cars3D is synthetic and other two datasets are real datasets.}
  \label{fig:data}
\end{figure}

\begin{table*}[t]
\begin{center}
\renewcommand{\arraystretch}{1.0}
\setlength\tabcolsep{10.5pt}

\begin{tabular}{lccccccc}

& & \multicolumn{3}{c}{LeNet} & \multicolumn{3}{c}{ResNet} \\ 
\toprule

 Dataset  & Approach 
 & Ref($\%$) & mAP(\%$\uparrow$) & Forget(\%$\downarrow$) 
 & Ref($\%$) & mAP(\%$\uparrow$) & Forget(\%$\downarrow$) \\
\midrule

   \multirow{5}{*}{Cars3D}  
& Naïve & $68.69$ & $46.28$ & $32.89$ & $70.82$ & $50.01$ & $29.38$ \\
&  FT & $68.69$ & $\textbf{49.36}$ & $\textbf{28.43}$ & $70.82$ & $44.98$ & $36.49$ \\
&  LFL & $68.69$ & $33.05$ & $52.08$ & $70.82$ & $33.96$ & $52.05$  \\
&  LwF & $68.69$ & $-$ & $-$ &  $70.82$ & $-$ & $-$ \\
&  EWC & $68.69$ & $46.19$ & $33.02$ &$70.82$ & \textbf{50.91} & \textbf{28.11} \\

\midrule

   \multirow{5}{*}{MVCD}  
& Naïve & $83.22$ & $62.31$ & $25.13$ & $94.99$ & \textbf{83.77} & \textbf{11.81}\\
& FT & $83.22$ & $60.65$ & $27.13$ & $94.99$ & $72.75$ & $23.41$ \\
&  LFL & $83.22$ & $\textbf{62.86}$ & $\textbf{24.47}$ & $94.99$  & $66.24$ & $30.27$  \\
&  LwF & $83.22$ & $-$ & $-$ & $94.99$ & $-$ & $-$ \\
&  EwC & $83.22$& $61.44$ & $26.18$ & $94.99$ & $81.44$ & $14.26$ \\
\midrule 

   \multirow{5}{*}{CompCars} & Naïve & $29.83$ & $\textbf{18.25}$ & $\textbf{38.82}$ & $54.41$ & $\textbf{38.60}$ & $\textbf{29.06}$\\

&  FT & $29.83$ & $12.51$ & $58.06$ & $54.41$ &  $22.63$ & $58.41$ \\
&  LFL & $29.83$ & $11.18$ & $62.52$ & $54.41$ & $25.24$ & $53.61$\\
&  LwF & $29.83$ & $-$ & $-$ & $54.41$ & $-$ & $-$ \\\
&  EWC & $29.83$ & $17.19$ & $42.37$ & $54.41$ & $37.43$ & $31.21$\\

\midrule 

\end{tabular}
   \caption{Benchmarking existing techniques.}
\label{tab:benchmark}
\end{center}
\end{table*}

\subsection{Datasets} 

We use three publicly available datasets throughout our experiments. For each dataset, we split the training data in a continuous incremental batch fashion. We do not make use of any dataset-specific procedure (\ie augmentation) in our experiments.

\partitle{\textbf{Cars3D}} \cite{Cars3D} Consists of 183 car models, for each model we render 96 data points generated from 24 equally spaced azimuth directions and 4 elevations. We use 100 car models for training. For evaluating, we use 83 car models. From each, we randomly select 10 images for the query set and 86 images for the gallery set. In total, the training set consists of 9600 images, split into 10 incremental batches. The query and gallery set consist of 830 and 6972 images respectively.
    
 \partitle{\textbf{Multi-View Car Dataset}} \cite{CarsEPFL} (MVCD) Consists of a sequence of images from 20 different car models on a rotating platform. An image was taken every 3 to 4 degrees. For training, we select 15 car models, comprising of 1737 images split into 5 incremental batches. From the remaining 5 car models, we select 10 images for the query set and the remaining for the gallery set.
    
 \partitle{\textbf{CompCars}} \cite{compcars} We use the surveillance-nature subset which contains 50,000 car images, captured only from the frontal view. In total, there are 281 different car models. The number of images for each model ranges from 50 to 500 different captures. We use 240 car models for our training set, a total of 36737 images split into 10 incremental batches. For the query set, we randomly select 20 images of the unseen cars and use the remaining images as the gallery set.


\subsection{Backbone Architectures} 
We use two different backbone architectures - both trained from scratch. All models are trained until convergence. We first train a model with the first continuous batch and then re-train the same model with subsequent continuous batches and a continual instance learning approach. We repeat this process until all instances are trained.

Unless specifically stated, we use the same
training procedure across all experiments and
datasets. We implement our experiments using the PyTorch framework \cite{pytorch}. We use the Adam optimizer \cite{Adam} with $\beta_1$ = $0.9$, $\beta_2$ = $0.999$, 
$\epsilon$ = $10^{-3}$ for the LeNet experiments and $\epsilon$ = $10^{-4}$ for the ResNet experiments.

\vspace{\baselineskip}

\partitle{LeNet} - We use a LeNet-5 \cite{LeNet} like architecture with 3 convolutional layers and 3 fully connected layers. We apply batch normalisation \cite{batch_norm}, ReLU non-linearities and 2 \texttt{x} 2 max-poolings with stride 2. 
All layers are initialised with Xavier uniform \cite{xavier_init}. For all datasets, we train the network with the 32 \texttt{x} 32 RGB images and define the last fully connected layer with $D$-dimension 32.

\vspace{\baselineskip}

\partitle{ResNet} - We employ the ResNet-18 \cite{resnet} with 64 \texttt{x} 64 RGB images for the Cars3D \cite{Cars3D} and the Multi-View Car Dataset \cite{CarsEPFL}, and 128 \texttt{x} 128 RGB images for the CompCars dataset \cite{compcars}. We use the last fully connected layer as the metric extractor with dimension 128.



\subsection{Evaluation.} 
In the re-identification evaluation setup, we
have a query set and a gallery set. For each vehicle in the query set, we aim to retrieve the same vehicle instances from the gallery set. We use mean-average-precision (mAP) for evaluating and comparing our approaches. The mAP is computed as the Average Precision (AP) across all queries. 

Under the continual learning framework, we employ a similar approach as~\cite{core50} and compare the relative mAP w.r.t. the corresponding cumulative approach - training with all instances - which is hereby referred to as \textit{Forget}. Our metrics on a Full Test Set: the query and gallery sets remain fixed throughout the continuous batches~\cite{core50}.

\begin{table*}[t]
\begin{center}
\renewcommand{\arraystretch}{1.0}
\setlength\tabcolsep{6pt}

\begin{tabular}{lccccc}

& & \multicolumn{2}{c}{LeNet} & \multicolumn{2}{c}{ResNet} \\ 
\toprule

 Dataset  & Approach 
 & Regression(\%$\downarrow$) & w/ NCE(\%$\downarrow$)
 & Regression(\%$\downarrow$) & w/ NCE(\%$\downarrow$)  \\
\midrule

  \multirow{5}{*}{Cars3D}  
& Naïve & $32.89$ & $25.64$ & $29.38$ & $30.61$ \\
&  FT & $28.43$ & $24.02$  &  $36.49$ & $33.71$ \\
&  LFL & $52.08$ & $33.50$  &$52.05$ & $43.10$ \\
&  LwF & $-$ & $\textbf{18.29}$ & $-$ & $30.19$ \\
&  EWC & $33.02$ & $23.66$  & $28.11$ & $\textbf{19.87}$  \\

\midrule 

  \multirow{5}{*}{MVCD}  
& Naïve & $25.13$ & $22.80$ & $\textbf{11.81}$ & $16.10$  \\
& FT & $27.13$ & $\textbf{20.37}$ & $23.41$ & $24.16$  \\
&  LFL &$24.47$ & $20.76$ & $30.27$ & $27.11$ \\
&  LwF & $-$ & $21.36$ & $-$ & $14.59$  \\
&  EWC & $26.18$ & $28.44$ & $14.26$ & $13.92$ \\

\midrule 

  \multirow{5}{*}{CompCars}
  & Naïve & $38.82$ & $35.13$ & $\textbf{29.06}$ & $39.77$ \\

&  FT & $58.06$ &  $51.66$ & $58.41$ & $52.38$ \\
&  LFL & $62.52$ & $58.26$ & $53.61$ & $54.47$ \\
&  LwF & $-$ & $49.55$ & $-$ & $60.37$ \\
&  EWC & $42.37$ & $\textbf{32.28}$ & $31.21$ & $29.13$ \\

\midrule 

\end{tabular}
  \caption{Contribution of Normalised Cross Entropy (NCE) to Continual Instance Learning - Forget ratio.}
\label{tab:NCE}
\end{center}

\end{table*}

\section{Experiments}

Our experiments aim at answering the following questions \textbf{Q1)} What is the performance of existing continual instance learning methods for object instance recognition?
\textbf{Q2)} Can Normalised Cross-Entropy improve CIL?
\textbf{Q3)} Can synthetic transfer learning improve CIL?

\subsection{Q1: Performance of Existing Techniques}

In Table \ref{tab:benchmark}, we evaluate the different continual learning approaches under the CIL setup. Additionally, we provide the cumulative mAP as an offline reference. Similar to the continual classification task, CIL also suffers from catastrophic forgetting.

In general, the Naïve approach outperforms others. Additionally, EWC approach also has good performance. In both approaches, the network is allowed to fully improve its decision boundary and its feature extraction capacity. The LFL approach has, in general, the lowest performance. We hypothesize that keeping the embedding unchanged is not desirable when we aim to learn the embedding itself. The ResNet-18 model performs better under the cumulative approach and suffers from less forgetting when compared to the LeNet model. The CompCars dataset is the most challenging set under offline and CIL, whereas the MVCD is the easiest.


In this experiment, we conclude that Continual Instance Learning is a challenging problem for current CL approaches, and the existing methods perform comparably or inferior to the Naïve approach.

\subsection{Q2: Contribution of NCE}

In our second experiment, we investigate the effect of NCE, Table~\ref{tab:NCE}. In all experiments, we notice that the more negatives samples we add, the better the results. We restrict to nine negative samples.



We identify that most methods benefit from NCE. This is particularly true for EWC, where the feature extracting layers benefit the most from bounding the gradients. Overall, both architectures gain from NCE. This is more visible in the LeNet model which is more prone to over-fitting and more highly impacted by regression outliers.

Integrating NCE to CIL is straightforward and beneficial to most approaches.

\begin{table*}[t]
\begin{center}
\renewcommand{\arraystretch}{1.0}
\setlength\tabcolsep{4.5pt}

\begin{tabular}{lcccccc}

& & \multicolumn{2}{c}{LeNet} & \multicolumn{2}{c}{ResNet} \\ 
\toprule

 Dataset  & Approach 
 & Regression($\%\downarrow$)  & w/ NCE+Transfer(\%$\downarrow$)
 & Regression($\%\downarrow$)  & w/ NCE+Transfer(\%$\downarrow$) \\
\midrule

  \multirow{5}{*}{MVCD}  
& Naïve & $25.13$  & $9.84$ & $11.81$ &  $\textbf{0.48}$ \\
& FT & $27.13$  & $10.15$ & $23.41$ &  $19.23$\\
&  LFL &$24.47$  & $14.25$ & $30.27$ &  $13.16$\\
&  LwF & $-$  &  $\textbf{3.08}$ & $-$ &  $12.12$\\
&  EWC & $26.18$  & $9.22$ & $14.26$ &  $2.29$\\

\midrule 

  \multirow{5}{*}{CompCars} 
  & Naïve & $38.82$ & $\textbf{37.85}$ & $\textbf{29.06}$ &  $33.82$\\

&  FT & $58.06$  & $71.74$ & $58.41$ &  $70.99$\\
&  LFL & $62.52$  & $61.28$ & $53.61$ &  $50.01$\\
&  LwF & $-$  & $60.44$ & $-$ &  $41.15$\\
&  EWC & $42.37$  & $39.73$ & $31.21$ &  $31.35$\\

\midrule 

\end{tabular}
  \caption{Contribution of Synthetic Transfer to Continual Instance Learning - Forget ratio.}
\label{tab:transfer}
\end{center}

\end{table*}


\subsection{Q3: Contribution of Synthetic Transfer}

In Table~\ref{tab:transfer}, we investigate the effects of pre-training a network on synthetic data on CIL approaches. Synthetic object instances can be found in publicly available datasets \cite{shapenet, objectnet}. Specifically, in our experiments, we pre-train a model on the entire Cars3D dataset and perform continuous batch training on MVCD and CompCars. We reduced the learning rate to $10^{-4}$ for all continual experiments and froze the convolutional layers. 


All methods, under both architectures, benefit from our method. Particularly, Naïve and EWC. This is more evident in MVCD. The gains are not as visible in CompCars. This is likely due to the differences in the data distributions. The synthetic dataset consists of car images in different azimuths and elevations. Conversely, CompCars has only frontal view images. This restricts the visual cues to features, which are limited in Cars3D, such as the brand, the frontal grill, and the headlamps. Overall, the ResNet model benefits the most from synthetic transfer.

Having a relevant distributed synthetic dataset can reduce forgetting in CIL in combination with NCE.

\section{Conclusion}







In this paper, we studied continual learning of object instances. Firstly, we verified that existing techniques suffer from catastrophic forgetting while learning object instances. We show that the simple Naïve approach performs competitively to other continual learning techniques.  
This indicates that existing methods are unsuitable for continual instance learning, calling for specific techniques. To that end, we incorporated normalised cross-entropy along with synthetic visual transfer to existing techniques to circumvent forgetting. We observed that indeed cross-entropy regulates the learning, and synthetic visual data is beneficial, especially when it is similar to the target data. From our observations, we foresee two plausible directions for future research. In this paper, we solely build upon regularisation approaches, however, replay-based techniques could be promising for continual instance learning. Secondly, our paper focused on car instances, and we leave the exploration of other rigid object instances, such as household objects, or nonrigid object instances such as humans or human faces as future work.



\section*{Acknowledgement}
We thank Prosus for partially funding this research and providing the computational resources. Also, we thank Dmitri Jarnikov, Samarth Bhargav and Manasa Bhat for the constructive feedback.

{\small
\bibliographystyle{ieee_fullname}
\bibliography{0-main}

\begin{thebibliography}{10}\itemsep=-1pt

\bibitem{siameseCNN}
Jane Bromley, James Bentz, Leon Bottou, Isabelle Guyon, Yann Lecun, Cliff
  Moore, Eduard Sackinger, and Rookpak Shah.
\newblock Signature verification using a "siamese" time delay neural network.
\newblock {\em IJPRAI}, 1993.

\bibitem{shapenet}
Angel~X. Chang, Thomas Funkhouser, Leonidas Guibas, Pat Hanrahan, Qixing Huang,
  Zimo Li, Silvio Savarese, Manolis Savva, Shuran Song, Hao Su, Jianxiong Xiao,
  Li Yi, and Fisher Yu.
\newblock {ShapeNet: An Information-Rich 3D Model Repository}.
\newblock Technical Report arXiv:1512.03012, 2015.

\bibitem{sim_clr}
Ting Chen, Simon Kornblith, Mohammad Norouzi, and Geoffrey Hinton.
\newblock A simple framework for contrastive learning of visual
  representations.
\newblock {\em arXiv preprint arXiv:2002.05709}, 2020.

\bibitem{face_veri}
Sumit Chopra, Raia Hadsell, and Yann LeCun.
\newblock Learning a similarity metric discriminatively, with application to
  face verification.
\newblock In {\em CVPR}, 2005.

\bibitem{pascal}
Mark Everingham, Luc Van~Gool, Christopher~KI Williams, John Winn, and Andrew
  Zisserman.
\newblock The pascal visual object classes (voc) challenge.
\newblock {\em IJCV}, 2010.

\bibitem{syntheticData}
Adrien Gaidon, Antonio López, and Florent Perronnin.
\newblock The reasonable effectiveness of synthetic visual data.
\newblock In {\em IJCV}, 2018.

\bibitem{xavier_init}
Xavier Glorot and Yoshua Bengio.
\newblock Understanding the difficulty of training deep feedforward neural
  networks.
\newblock In {\em AISTATS}, 2010.

\bibitem{empiricalCatastrophicForgetting}
Ian~J. Goodfellow, Mehdi Mirza, Da Xiao, Aaron Courville, and Yoshua Bengio.
\newblock An empirical investigation of catastrophic forgetting in
  gradient-based neural networks.
\newblock In {\em ICLR}, 2013.

\bibitem{contrastive_loss}
Raia Hadsell, Sumit Chopra, and Yann LeCun.
\newblock Dimensionality reduction by learning an invariant mapping.
\newblock In {\em CVPR}, 2006.

\bibitem{resnet}
Kaiming He, Xiangyu Zhang, Shaoqing Ren, and Jian Sun.
\newblock Deep residual learning for image recognition.
\newblock In {\em CVPR}, 2016.

\bibitem{InDefenseOfTripletLoss}
Alexander Hermans, Lucas Beyer, and Bastian Leibe.
\newblock In defense of the triplet loss for person re-identification.
\newblock In {\em CVPR}, 2017.

\bibitem{kd}
Geoffrey Hinton, Oriol Vinyals, and Jeff Dean.
\newblock Distilling the knowledge in a neural network.
\newblock In {\em NIPS}, 2014.

\bibitem{batch_norm}
Sergey Ioffe and Christian Szegedy.
\newblock Batch normalization: Accelerating deep network training by reducing
  internal covariate shift.
\newblock In {\em JMLR}, 2015.

\bibitem{lfl}
Heechul Jung, Jeongwoo Ju, Minju Jung, and Junmo Kim.
\newblock Less-forgetting learning in deep neural networks.
\newblock {\em arXiv}, 2016.

\bibitem{Adam}
Diederik Kingma and Jimmy Ba.
\newblock Adam: A method for stochastic optimization.
\newblock In {\em ICLR}, 2014.

\bibitem{kirkpatrick2017overcoming}
James Kirkpatrick, Razvan Pascanu, Neil Rabinowitz, Joel Veness, Guillaume
  Desjardins, Andrei~A Rusu, Kieran Milan, John Quan, Tiago Ramalho, Agnieszka
  Grabska-Barwinska, et~al.
\newblock Overcoming catastrophic forgetting in neural networks.
\newblock 2017.

\bibitem{imagenet}
Alex Krizhevsky, Ilya Sutskever, and Geoffrey~E Hinton.
\newblock Imagenet classification with deep convolutional neural networks.
\newblock In {\em NIPS}, 2012.

\bibitem{VehicleReID_Triplet}
Ratnesh Kuma, Edwin Weill, Farzin Aghdasi, and Parthasarathy Sriram.
\newblock Vehicle re-identification: an efficient baseline using triplet
  embedding.
\newblock In {\em IJCNN}, 2019.

\bibitem{LeNet}
Yann LeCun, L{\'e}on Bottou, Yoshua Bengio, and Patrick Haffner.
\newblock Gradient-based learning applied to document recognition.
\newblock {\em Proceedings of the IEEE}, 1998.

\bibitem{lwf}
Zhizhong Li and Derek Hoiem.
\newblock Learning without forgetting.
\newblock In {\em ECCV}, 2016.

\bibitem{mscoco}
Tsung-Yi Lin, Michael Maire, Serge Belongie, James Hays, Pietro Perona, Deva
  Ramanan, Piotr Doll{\'a}r, and C~Lawrence Zitnick.
\newblock Microsoft coco: Common objects in context.
\newblock In {\em ECCV}.

\bibitem{core50}
Vincenzo Lomonaco and Davide Maltoni.
\newblock Core50: a new dataset and benchmark for continuous object
  recognition.
\newblock {\em PMLR}, 2017.

\bibitem{mclelland}
James Mcclelland, Bruce Mcnaughton, and Randall O’Reilly.
\newblock Why there are complementary learning systems in the hippocampus and
  neocortex: Insights from the successes and failures of connectionist models
  of learning and memory.
\newblock {\em Psychological review}, 1995.

\bibitem{McCloskey_Cohen_1989}
Michael McCloskey and Neal~J Cohen.
\newblock Catastrophic interference in connectionist networks: The sequential
  learning problem.
\newblock In {\em Psychology of learning and motivation}. 1989.

\bibitem{CarsEPFL}
Mustafa Ozuysal, Vincent Lepetit, and Pascal Fua.
\newblock Pose estimation for category specific multiview object localization.
\newblock In {\em CVPR}, 2009.

\bibitem{less_labels}
Johannes~C Paetzold, Oliver Schoppe, Rami Al-Maskari, Giles Tetteh, Velizar
  Efremov, Mihail~I Todorov, Ruiyao Cai, Hongcheng Mai, Zhouyi Rong, Ali
  Ertuerk, et~al.
\newblock Transfer learning from synthetic data reduces need for labels to
  segment brain vasculature and neural pathways in 3d.
\newblock In {\em MIDL}, 2019.

\bibitem{pytorch}
Adam Paszke, Sam Gross, Soumith Chintala, Gregory Chanan, Edward Yang, Zachary
  DeVito, Zeming Lin, Alban Desmaison, Luca Antiga, and Adam Lerer.
\newblock Automatic differentiation in pytorch.
\newblock In {\em NIPS}, 2017.

\bibitem{Cars3D}
Scott~E Reed, Yi Zhang, Yuting Zhang, and Honglak Lee.
\newblock Deep visual analogy-making.
\newblock In {\em NIPS}. 2015.

\bibitem{humanpose_synthetic}
Gr{\'e}gory Rogez and Cordelia Schmid.
\newblock Image-based synthesis for deep 3d human pose estimation.
\newblock In {\em IJCV}, 2018.

\bibitem{synthetic_fog}
Christos Sakaridis, Dengxin Dai, and Luc Van~Gool.
\newblock Semantic foggy scene understanding with synthetic data.
\newblock In {\em IJCV}, 2017.

\bibitem{facenet}
Florian Schroff, Dmitry Kalenichenko, and James Philbin.
\newblock Facenet: A unified embedding for face recognition and clustering.
\newblock In {\em CVPR}, 2015.

\bibitem{NML}
Kihyuk Sohn.
\newblock Improved deep metric learning with multi-class n-pair loss objective.
\newblock In {\em NIPS}, 2016.

\bibitem{incremental_learning}
Stefan Stojanov, Samarth Mishra, Ngoc~Anh Thai, Nikhil Dhanda, Ahmad Humayun,
  Chen Yu, Linda~B. Smith, and James~M. Rehg.
\newblock Incremental object learning from contiguous views.
\newblock In {\em CVPR}, June 2019.

\bibitem{objectnet}
Yu Xiang, Wonhui Kim, Wei Chen, Jingwei Ji, Christopher Choy, Hao Su, Roozbeh
  Mottaghi, Leonidas Guibas, and Silvio Savarese.
\newblock Objectnet3d: A large scale database for 3d object recognition.
\newblock In {\em ECCV}, 2016.

\bibitem{compcars}
Linjie Yang, Ping Luo, Chen~Change Loy, and Xiaoou Tang.
\newblock A large-scale car dataset for fine-grained categorization and
  verification.
\newblock In {\em CVPR}, 2015.

\bibitem{duplicate}
Yi Zhang, Yanning Zhang, Jinqiu Sun, Haisen Li, and Yu Zhu.
\newblock Learning near duplicate image pairs using convolutional neural
  networks.
\newblock 2018.

\end{thebibliography}
}

\end{document}